\def\year{2019}\relax
\documentclass[letterpaper]{article} 
\usepackage{aaai19}  
\usepackage{times}  
\usepackage{helvet} 
\usepackage{courier}  
\usepackage[hyphens]{url}  
\usepackage{graphicx} 
\def\UrlFont{\rm}  
\usepackage{graphicx}  
\usepackage[alpha]{mdpn}
\usepackage{import}
\usepackage{tikz} 

\usepackage{layers/Ball}
\usepackage{layers/Box}
\usepackage{layers/RightBandedBox}
\usepackage{algorithm}
\usepackage{algpseudocode}
\algdef{SE}[DOWHILE]{Do}{doWhile}{\algorithmicdo}[1]{\algorithmicwhile\ #1}%

\title{Improving Deep Reinforcement Learning in Minecraft with Action Advice}

\author{ \\ \Large \textbf{Spencer Frazier, Mark Riedl}\\ 
Georgia Institute of Technology - Human-Centered AI Lab\\
}

\nocopyright

\begin{document}

\maketitle

\begin{abstract}
Training deep reinforcement learning agents complex behaviors in 3D virtual environments requires significant computational resources.
This is especially true in environments with high degrees of aliasing, where many states share nearly identical visual features. Minecraft is an exemplar of such an environment.
We hypothesize that {\em interactive machine learning} (IML), wherein human teachers play a direct role in training through demonstrations, critique, or action advice, may alleviate agent susceptibility to aliasing.
However, interactive machine learning is only practical when the number of human interactions is limited, requiring a balance between human teacher effort and agent performance.
We conduct experiments with two reinforcement learning algorithms which enable human teachers to give action advice---Feedback Arbitration, and Newtonian Action Advice---under visual aliasing conditions.
To assess potential cognitive load per advice type, we vary the accuracy and frequency of various human action advice techniques. The training efficiency, robustness against infrequent and inaccurate advisor input, and sensitivity to aliasing are examined.

\end{abstract}

\section{Introduction}

\begin{figure}[t]
\centering
\includegraphics[width=0.75\columnwidth]{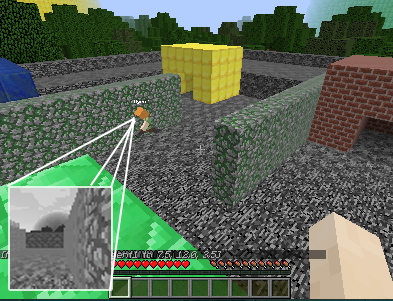} 
\caption{A questing environment in Minecraft. The agent's view is shown in the inlay.}
\label{fig:minecraft}
\end{figure}

\begin{figure}[t]
\centering{
\includegraphics[width=0.75\columnwidth]{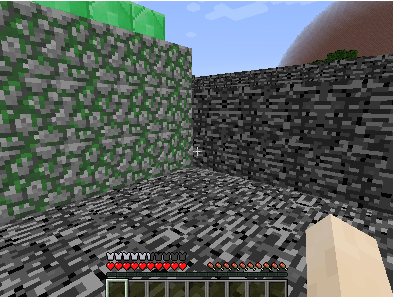}} 
\caption{An example of perceptual aliasing in Minecraft.
}
\label{fig:aliasing}
\end{figure}

Training deep reinforcement learning agents requires significant time and computational resources. This is especially true in rich, virtual, 3D dimensional environments with large state spaces such as the game, Minecraft (See Figure~\ref{fig:minecraft}). Latent states must be inferred from noisy, pixel-level data and high amounts of perceptual aliasing confound interpretation of the underlying state. 
{\em Perceptual aliasing} occurs when many states share nearly identical visual features. 
Figure~\ref{fig:aliasing} shows an example of an area with a lot of perceptual aliasing.
A consequence of perceptual aliasing is that deep reinforcement learning agents struggle to learn the relationship between visual features and the utility of particular actions when those features are present---sometimes one action must be taken and sometimes another action must be taken. 
This can cause the agent to fail to converge on a policy and/or to require additional exploration of areas of high aliasing to learn better feature representations.
Many environments, including for games and those used in machine learning research (c.f., \cite{zhiyu}) seed an environment with distinguishing features, such as different wall colors or structural markers. 
These interventions in the environment to aide agent learning are artificial.
Because of these factors, it is non-trivial to train a reinforcement learning agent to perform optimally in increasingly realistic environments with high amounts of perceptual aliasing. 

Human intervention can help speed up reinforcement learning by assisting a deep reinforcement learning (DRL) agent when its confidence in its learned policy is low.
{\em Interactive machine learning} (IML) \cite{iml} seeks to improve upon traditional machine learning algorithms by allowing humans to play a direct role in training. Human collaborators provide demonstrations of correct behavior, providing action advice before a policy action is taken, or providing critique after the agent acts.
Demonstrations are very effective but can often be costly, difficult, or time-consuming to produce and record.
Action advice and critique allow a human teacher to monitor the agent's learning from remote and provide feedback that biases it to better learning outcomes.
Studies \cite{krening} show that non-experts prefer giving action advice over critique. 
However, a majority of studies on advice and critique giving are conducted in grid-world environments. An agent must navigate through a 2D grid in which the agent's $(x,y)$ location is a given feature.
3D virtual environments add a layer of complexity by requiring an agent to learn features from pixel-level input that correspond to utility toward a goal and
make agents susceptible to perceptual aliasing.

In this paper, we attempt to understand how Learning from Advice (LfA) algorithms perform in 3D environments with perceptual aliasing.
We adapt two LfA algorithms to the Minecraft environment and train them to perform an item delivery quest in a map with high perceptual aliasing.
The {\em Feedback Arbitration}~\cite{zhiyu} algorithm is a straightforward technique for incorporating human action advice into reinforcement learning in which the agent accepts action advice whenever it is provided by a human teacher and the agent has low confidence in its own policy.
The {\em Newtonian Action Advice} algorithm accepts human action advice whenever provided and assumes any piece of advice remains applicable for a period of time afterward even if no other advice is given.

Effective IML requires balancing the learning performance of the agent against the cognitive load demand on the human teacher.
The teacher may give intermittent advice because they are fatigued or distracted.
Similarly, error, confusion, or distraction may degrade advice accuracy.
In order to run controlled studies assessing different rates of teacher response and teacher accuracy, we construct an {\em oracle}, a synthetic teacher with perfect environment information. 
The oracle can be parameterized with regard to (a)~how often to provide action advice to a reinforcement learning agent and (b)~how often to give incorrect advice. 

Our contributions show that high frequency advice is superior to low frequency advice and greatly superior to baseline reinforcement learning. Further, we show that even significantly inaccurate advice is better than no advice. 
%
Experimental results show that Newtonian Action Advice uses advice feedback more efficiently allowing it to mitigate time spent in aliased states.


\section{Background and Related Work}

Reinforcement Learning is proven to be effective at learning agent control policies in 2D and 3D game environments including Atari~\cite{drqn}, navigation in Minecraft environments~\cite{liu,oh,singh}, Doom~\cite{doom}, Quake II~\cite{ctf}, DOTA 2, and StarCraft II.
Most game environments are designed to have low perceptual aliasing to make it easier on human players even though the real world can often have high perceptual aliasing.
3D computer game environments provide a stepping stone to application of reinforcement learning in real world environments~\cite{laird,ctf} but sometimes overlook this aspect of the real world.

Perceptual aliasing is objectively higher when an agent is closely facing a wall due to repeated block textures in Minecraft.
Walls and structure boundaries are exemplars for aliased states because it guarantees the texture is repeated elsewhere in the maze. 
Walls and structures with significant height can also block the agent's view of distant but unique visual landmarks that it might otherwise learn to orient itself.
Another example of perceptual aliasing occurs in longer stretches of maze-like environments.
To a mazerunner, every point along the path between two hedges, a corridor, or a tunnel can look similar.
(Figure~\ref{fig:minecraft} shows a short hedge maze).
Often there are non-perceptual features that afford better task performance in these situations that reinforcement learning agents don't explicitly model, such as length of time traveling straight.
Conventional deep reinforcement learning algorithms must spend more time exploring high-aliasing parts of the environment.

Human feedback in IML can take different forms. 
{\em Learning from Demonstration} (LfD) allows humans to directly provide examples of proper behavior \cite{argall}. 
From demonstrations, an agent can learn the policy directly, learn to explore more effectively \cite{griffith}, or learn a reward function from which to reconstruct a policy \cite{abbeel}. 
%
However, it is not always feasible to provide demonstrations. 
{\em Learning from Critique} (LfC) allows human teachers to indicate that the agent is doing well or not doing well in order to bias the agent toward certain outcomes. Learning from Critique can also include human indication of preferences over variations in agent behavior \cite{critique}. 
{\em Learning from Advice} (LfA) is similar to Learning from Critique, except the human teacher proactively advises the agent on the actions it should take instead of retroactively rewarding or punishing the agent for an action it already took.
Since studies show that non-expert human teachers prefer giving action advice over critique~\cite{krening} we focus on Learning from Action. 

We build upon a rich set of research exploring human feedback in reinforcement learning~\cite{griffith,krening,thomaz,isbell,stone,littman}. A majority of this work assesses the algorithms in 2D grid-world environments where the agent’s $(x,y)$ location is a given feature. Some do address 3D \cite{Imperfect}, more simplistic 3D environments \cite{zhiyu}, or augment the agent with non-visual state information \cite{abel}. Overall there are still open questions about the scalability of these algorithms in 3D environments.


We build off two LfA algorithms in our exploration of perceptual aliasing in 3D virtual environments.
The first, {\em Feedback Arbitration}~\cite{zhiyu} is based on a standard deep reinforcement learning technique called {\em Deep $Q$ Networks} (DQN).
Deep $Q$ Networks \cite{mnih2} use a convolutional neural network to learn visual features in the state that correspond to the utility of different actions---called a $q$-value.
A Feedback Arbitration agent alternates between exploring the environment, exploiting it's $Q$ network, and listening to a human teacher depending on (a) its confidence in its own $Q$ network and (b) it's learned confidence in the human trainer.
Feedback Arbitration was tested in very simple 3D grid-world environment with landmarking features which helped lower perceptual aliasing.

The second, the {\em Newtonian Action Advice} algorithm~\cite{krening} is built on tabular $Q$-learning, where a table of $q$-values is stored for every state and action. Krening et al.~\shortcite{KreningSubjective} found that human trainers perceived a Newtonian Action Advice agent as more intelligent, more transparent, more performant and less frustrating than a standard $q$-learning agent.
The standard $Q$-learning algorithm is enhanced such that if a human trainer provides action advice the agent takes the advice depending on its confidence in its $q$-table.  
If there is no action advice, the agent continues to follow the last received advice for $t$ additional timesteps. 
There is no neural component to NAA and requires an enumerable state space; as such it is not directly scalable to 3D environments.
We extend this approach to work with a DQN instead of a $q$-table, allowing it to operate in 3D graphical environments.




\section{The Minecraft Environment}


Minecraft is a 3D game in which players can move, explore and build. 
Minecraft allows for continuous movement, is partially observable, and is mostly deterministic and dynamic.
While Minecraft uses a graphical aesthetic of blocks and thus simpler than photo-realistic graphical environments, it easily accommodates large, easily modifiable, open worlds. 
This makes Minecraft a valuable stepping stone to more real environments;
other 3D game environments tend to have smaller state spaces.
{\em Project Malmo} \cite{malmo} provides an API for reinforcement learning agents to control virtual characters within a Minecraft map.

\begin{figure}[t]
\centering
\includegraphics[width=0.65\columnwidth]{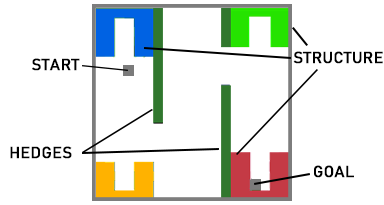} 
\caption{Environment map with 3D structure and key points labeled.}
\label{fig:map}
\end{figure}

Whereas many games and many research projects using Minecraft seek to reduce perceptual aliasing, we designed a 20x20 Minecraft level to have a mixture of high perceptual aliasing and low perceptual aliasing. 
See Figure~\ref{fig:minecraft}.
High walls around the border of the map and a hedge corridor create areas of high perceptual aliasing. 
Three colored buildings and a stationary non-player character provide areas of low perceptual aliasing.
The map is mirrored across the x-axis, the center area is identical in each cardinal direction and the majority of surfaces share repeating textures.
The map schematic is shown in Figure~\ref{fig:map}.

The agent must interact with the environment to learn a policy to optimally reach the non-player character on the far side of the map.
The task approximates the common situation found in many 3D role-playing games where the player must fund a non-player character to deliver an item or receive a quest. 

For purposes of this paper, only 4 discrete actions are permitted: move forward, turn left, turn right and a complete 180-degree turn. 
%
The agent reward signal is as follows. 
The agent receives $-0.5$ points for each action taken and $+15000$ for reaching the non-player character on the far side of the map.
There are two intermediate rewards of 
 $+1500$ given when the agent reaches a $x$-coordinate corresponding to the entrance of the hedge corridor and when the agent reaches an $x$-coordinate corresponding to the exit from the hedge corridor.

%




\section{Reinforcement Learning Preliminaries}


A Markov Decision Process \cite{SuttonBarto}, MDP, is a model used to describe a potentially stochastic sequential decision making problem. 
An MDP can be expressed as a tuple $\langle S,A,P,R,\gamma\rangle$ which contains:
\begin{itemize}
\item A set of possible world states $\sset$
\item A set of possible agent actions $\aset$
\item A transition function $\Tdef$ expressing the probability $Pr(s_{t+1} \mid s_{t}, a_{t})$ of transitioning to state $s_{t+1}$ from state $s_t$ when action $a_t$ is performed.  
\item A reward function $R(s,a): \sset \bigtimes \aset \rightarrow \mathbb{R}$
\item A discount factor $\gamma \in [0,1]$ that prioritizes immediate rewards over longer-term rewards
\end{itemize}

The solution to a MDP or POMDP is a policy $\pi:\sset \rightarrow a$, a function that dictates the best action an agent can take in any world state in order to maximize future rewards. 
Reinforcement Learning is a technique that learns an optimal policy for a MDP
when the transition function is not given.

$Q$-Learning \cite{qlearning} is an approach to reinforcement learning in which an agent uses trial-and-error in a simulated environment to converge on the utilities of performing certain actions in certain states, $Q(s,a)$.
$Q(s,a)$ is iteratively updated in each learning phase as follows:
\begin{equation}
\label{bellman}
\begin{split}
Q(s_t,a_t)=&Q(s_t,a_t)+\\
& \alpha[R(s_{t+1})+\gamma\underset{a'}{\max}\: Q(s_{t+1},a')-Q(s,a)]
\end{split}
\end{equation}
where $t$ is the previous time step and $\alpha$ is the learning rate.
The agent's policy is thus $a=argmax_{a'} Q(s,a')$.

For environments in which the states cannot be enumerated and maintaining a $q$-table is not possible, Deep Q-Learning \cite{dql} learns a deep neural network to approximate $Q(s,a)$.
This is typically done by training a multi-layered convolutional neural network (CNN) to identify features that are predictive of state-action utilities. This is also known as a deep $Q$ network (DQN) \cite{mnih2}. 
To train a DQN, the neural network attempts to predict $Q(s,a)$ given the current pixel array and the action just performed; the loss is computed as the difference between the current prediction and the one-step Bellman update in Equation~\ref{bellman}.

\section{Methods}


Two action-advice augmented reinforcement learning algorithms are used.
The first is based on the {\em Feedback Arbitration} (FA) algorithm~\cite{zhiyu}. 
An addition to the standard reinforcement learning training loop, an {\em arbiter}, continually assesses feedback consistency and quality versus the confidence it has in its learned policy.
Feedback Arbitration is already designed to work in 3D virtual environments similar to Minecraft though on much simpler maps.
The second is the {\em Newtonian Action Advice} (NAA) algorithm~\cite{krening}. 
Newtonian Action Advice was mainly assessed in 2D environments and uses tabular $q$-learning. 
To focus on the effects of perceptual aliasing, the mechanism for measuring human trainer consistency in FA is not used; only the confidence check 
is implemented. 
These algorithms and our modifications for perceptual aliasing experiments are described in further detail in the next sections.

A synthetic oracle is used in place of a human to overcome large sample size requirements, advice frequency variability and accuracy variability.
These dimensions cannot be controlled for using human trainers, however see \cite{krening} for human-subject experiments in grid worlds. 
The synthetic oracle has ``perfect'' knowledge of maze environment and always gives optimal advice unless explicitly tuned to have lower accuracy, in which case it gives random advice some percentage of the time. 
Another parameter determines how often the oracle provides advice.

\subsection{The Feedback Arbitration Agent}

The Feedback Arbitration (FA) agent is a standard DQN agent with the following updates. 
We assume that the oracle provides advice intermittently. 
Action advice is queued up in a pending advice array. Queued advice instances dequeue after some time.
The agent either (a)~explores using a random action, (b)~exploits its $Q$-network by picking the action it believes has the highest utility in the current state, or (c)~consults the pending advice array.
The agent chooses a random action according to the standard $\epsilon$-greedy strategy.
Otherwise, it measures its confidence in its $Q$-network and, if the confidence is low, chooses to consult the action advice instead.
Then the $Q$-network confidence score is computed as a cost function such that low confidence incurs a high cost:

\smallskip
\begin{equation}
relativeCost = \frac{-1}
     {\ln\sqrt(\frac{\min_{a\in A(s)} L_a}
     {L_{\mathtt{max}}})-1}
\end{equation}
\smallskip

\noindent
where $L_a$ is the loss value for the predicted activation of action $a$ in the current state  and $L_{\mathtt{max}}$ is the highest loss observed by the DQN thus far.
If $relativeCost(l) <= 0.25$ the agent acts using policy output. 
Otherwise, the agent utilizes the pending action advice.


\subsection{The Newtonian Action Advice Agent}

Newtonian Action Advice (NAA) relies on the same arbitration mechanism as FA but adds one key feature. 
Instead of atomic advice events, NAA persists advice input across timesteps. 
A ``friction''  
parameter specifies a count of timesteps 
during which the action advice is recited by the agent back to itself.
It is shown \cite{krening} that methods which persist human feedback can significantly reduce the cognitive load on human collaborators. 

The original implementation of NAA implements a confidence decay function for discounting advice as a function of time since it was given.
That decay function was not replicated herein. 

NAA was originally designed for grid-worlds where there is a one-to-one correspondence between actions and advice.
In a grid world advice is given as {\em up}, {\em down}, {\em left}, or {\em right} and that exactly corresponds to the actions the agent can perform. 
In Minecraft, the agent's actions are {\em walk forward}, {\em turn left}, {\em turn right}, and {\em turn around}.
Newtonian action advice does not make sense for turn actions---it rarely makes sense to continue turning once advised to do it once.
In our implementation of NAA, advice is provided in the form of movement in the cardinal directions: {\em north}, {\em south}, {\em east}, or {\em west}.
The agent converts this to instructions relative to its orientation; 
if the agent is not oriented to face the cardinal direction of the movement action advice, it turns to that orientation and then moves forward for $friction$ time steps.






\section{Experiments}

A deep $Q$-network architecture is shared by the Feedback Arbitration and Newtonian Action Advice agents. 
A four-layer convolutional neural network consumes state $s$ as a matrix consisting of the 50 most recent frames of the game. 
Pixel values are single channel (greyscale) to constrain model memory requirements. 
After each convolutional layer we employ a $3x3$ kernel, $relu$ activation, and batch normalization. 
We add a max pooling layer after each batch normalization. 
Final dense layers output a probability distribution across 4 different action types. 
We set the discount factor (gamma) to 0.95. 
We use the Adam optimizer. 
We use RMS logarithmic error as the loss function.


Because human oracle behavior is hypothesized to vary significantly, we simulate oracle advice with a synthetic oracle that is parameterized along two dimensions:
\begin{itemize}
\item {\em Frequency of advice}, which can be high (advice given ~5\% of the time) or low (advice given ~1\% of the time)
\item {\em Oracle accuracy}, which can be high (100\% accurate) or low (50\% accuracy)
\end{itemize}
This establishes four experimental conditions: high-frequency high-accuracy (HFHA), high-frequency low-accuracy (HFLA), low-frequency high-accuracy (LFHA), and low-frequency low-accuracy (LFLA).
Our experiments thus pinpoint four possible points on two dimensions that a human oracle might inhabit. 
They were chosen to be extreme to make it easier to observe how our agent behaves under differing potential human oracle behaviors.
In addition, a baseline RL agent using the same DQN architecture is included.
The baseline RL agent ignores all oracle advice.



\begin{figure}[t]
\centering{
\includegraphics[width=0.9\columnwidth]{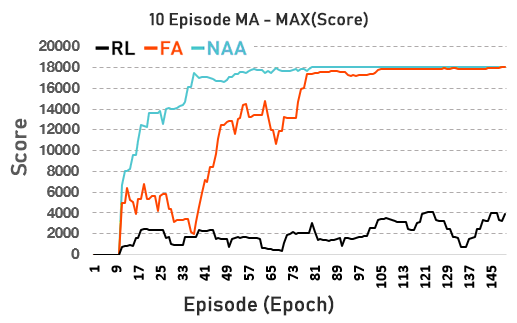}} 
\caption{Agent performance for all techniques as a 10-episode moving average of the max score across training runs across all 4 conditions.}
\label{fig:results-max}
\end{figure}

\begin{figure}[t]
\centering{
\includegraphics[width=0.9\columnwidth]{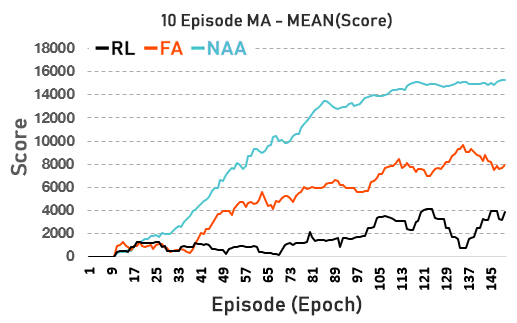}} 
\caption{Agent performance for all techniques as a 10-episode moving average of the mean score across training runs across all 4 conditions.}
\label{fig:results-mean}
\end{figure}

\begin{figure}[t]
\centering{
\includegraphics[width=0.9\columnwidth]{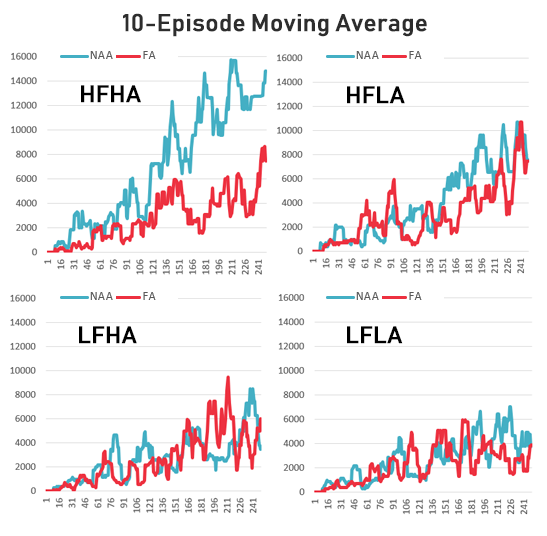}} 
\caption{Mean performance of FA vs NAA when varying accuracy and frequency of advice.}
\label{fig:results-varied}
\end{figure}



For each episode, the agent terminates after 1,000 actions (turns inclusive), or when successfully reaching the objective.
Each agent (RL, FA, NAA) was trained for at least 250 episodes under each of our four synthetic oracle conditions. 
This was repeated for at least 5 sessions per technique. 

In initial testing, the baseline RL agent struggled to converge when the terminating action count was set to 1,000. 
We therefore allow the RL agent to terminate after 1,500 actions to compensate, receiving 50\% more observations than the other techniques. 

The NAA agent is given a very conservative friction parameter, $friction=2$.


\section{Results}


We measure agent performance as the score achieved, which is identical to reward received.
Figure~\ref{fig:results-max} shows the maximum reward achieved plotted against training episode.
Figure~\ref{fig:results-mean} shows the mean reward achieved plotted against training episode.
As expected, the FA and NAA agents converge much faster than the baseline RL.
NAA converges faster than FA. 
This latter result is likely due to the intuition built into NAA that human trainer advice is intended to be continuous unless superseded.




\begin{figure*}[t]
\centering{
\includegraphics[width=0.85\textwidth]{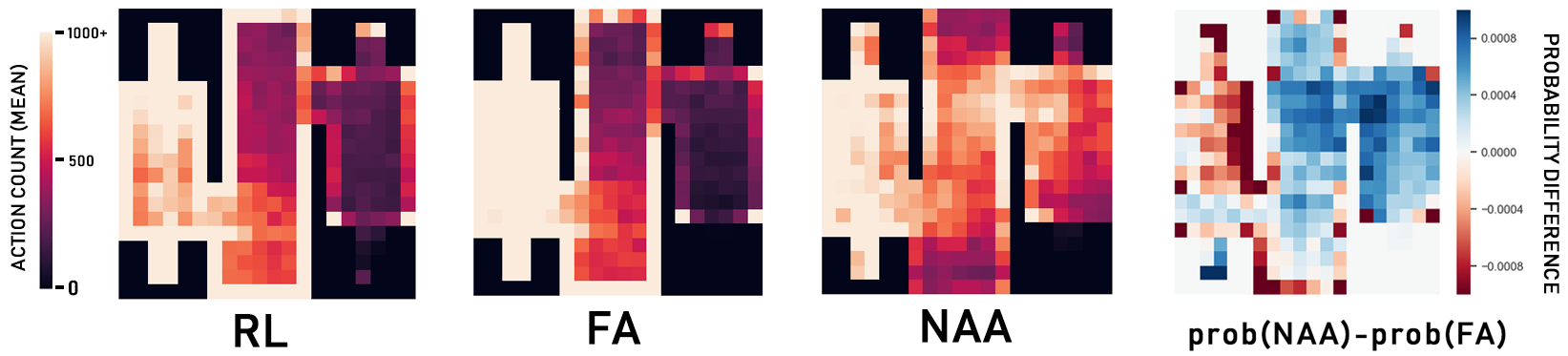}} 
\caption{Mean accumulated actions across techniques and the KL Divergence between FA and NAA.}
\label{fig:heat-map}
\end{figure*}

\subsection{Frequency and Accuracy Variation}

Four different conditions are established to assess FA and NAA sensitivity to varying advice frequency and oracle accuracy.
Results are shown in Figure~\ref{fig:results-varied}.
As expected, agents that receive high-accuracy advice outperform agents that receive low-accuracy advice. 
Likewise, agents that receive high-frequency advice outperform agents that receive advice less often.
NAA  dominates FA under the ideal circumstances of high advice frequency and high oracle accuracy.

No significant difference was observed in per-episode mean advice count for high-frequency (HF) conditions ($M_{NAA}$= 73, $M_{FA}$= 76). 
Per-episode, NAA exhibited a slightly higher reliance on advice across all conditions combined ($M_{NAA}$= 57, $M_{FA}$= 42).
Both agents received the same numbers of advice, but only access the advice when policy confidence is low.





One critique of reinforcement learning in games is that the policy is over specialized to a specific map.
We conducted an additional experiment in which we trained an agent for 250 episodes on the map in Figure~\ref{fig:map}---long enough for NAA and FA agents to converge.
We then continued training on the map rotated 90-degrees.
The NAA agent was able to re-converge in less than 40 episodes, demonstrating that filters within its $Q$-network were general enough to transfer.

\subsection{Robustness to Aliasing}

We hypothesize that Feedback Arbitration and Newtonian Action Advice will be less susceptible to perceptual aliasing in the environment. 
Our environment is specifically designed to have locations with high and low perceptual aliasing. 
High aliasing in particular happens around the walls of the map and in the hedge corridor in the middle.

We can inspect the different agents' robustness to aliasing by compiling a heatmap in which we count the number of times each agent visits a particular location in the map. 
In Figure~\ref{fig:heat-map} the first three heatmaps show location visit frequency for the baseline RL agent, FA agent, and NAA agent, respectively.
The baseline RL agent spends the most time near the start and in the first half of the hedge corridor, suggesting it does struggle with this area of high perceptual aliasing.
It also appears to trace along the walls of the hedge corridor, spending a lot of time in similar looking states.
The FA agent shows a similar pattern, though it does manage to converge on a policy whereas the baseline RL still struggles after hundreds of episodes. 

The heatmap for the NAA agent is qualitatively different.
Its visits to locations within the hedge corridor are much more uniform, suggesting that it navigates the high perceptual aliasing area with less difficulty. It also spends much less time fixated on walls.
Consequently, it spends a larger portion of its overall training time exploring locations near the goal. 
The right-most image shows the difference between the NAA agent's heatmap and the FA agent's heatmap;
the bluer the location, the more likely that location will be visited by the NAA agent instead of the FA agent.
The difference was computed using Kullback-Leibler Divergence between matrices containing the FA and NAA action counts.

\subsection{Discussion}

We conclude that Newtonian Action Advice improves agent learning performance inside the high perceptual aliasing areas.
Not only does the NAA agent converge on a policy that obtains high reward faster than the FA agent, but qualitatively the NAA agent spends a larger fraction of its training time moving decisively through areas of high perceptual aliasing.
This is achieved despite the fact that 
both FA and NAA receive the same advice, with the same oracle accuracy, and at the same frequency of advice. 
NAA is able to use this advice more effectively by assuming any advice remains valid for a period of time (2 timesteps in the experiments in this paper).
Coping with aliasing requires the agent leverage long-term memory to learn that it is okay to see nearly indistinguishable perceptual features for a number of time steps.
The oracle is acting as a memory aide until the learned policy overcomes confusion due to aliasing, which manifests itself as random exploration due to policy error.

The FA agent and NAA agent receive the same rate of advice from the synthetic oracle. 
However, the FA agent accessed the advice less often.
On first glance it seems FA might require less advice. 
However, we hypothesize that FA finds advice in high-aliasing situations less useful and is less reliant on oracle advice.
That is, it gains high confidence in its policy when tracing along the edges of the maze. Oracle advice to move into the middle of the hedge corridor does not help it disambiguate error in its $Q$-network.
The NAA agent, on the other hand, uses the given advice more often---confidence in its policy can be low due to aliasing though the recital of advice helps it cross highly aliased areas so that it ultimately converges on a robust policy.

\section{Conclusions}

Perceptual aliasing frequently occurs in real world environments, whereas computer games may often be designed to reduce aliasing. 
As demonstrated in this paper, it is relatively straightforward to create a task which is extremely difficult for a deep reinforcement learning agent to learn to solve. 
While we experiment with standard deep $Q$-networks, one can extrapolate that this will be true even for more sophisticated reinforcement learning algorithms. 
Learning from Advice (LfA) leverages the knowledge of a human teacher to overcome inefficiencies in learning due to perceptual aliasing. 
It is challenging to keep cognitive load low enough such that the human does not suffer from degraded human factors when interacting with the agent.
A human teacher may not be able to engage with the training agent and may give incorrect advice -- agents cant and must be robust to this.
We show that the Newtonian Action Advice LfA agent can be adapted to 3D environments and converges on a robust policy quickly despite the presence of perceptual aliasing, suggesting that it is using the oracle feedback more effectively than alternative techniques investigated.

 \section*{Acknowledgements}
 
 This material is based upon work supported by the Office of Naval Research (ONR) under Grant \#N00014-14-1-0003.

\end{document}